# Perturbated Gradients Updating within Unit Space for Deep Learning

Ching-Hsun, Tseng
Department of Computer Science
The University of Manchester
Manchester, UK
ching-hsun.tseng@postgrad.manchester.ac.uk

Hsueh-Cheng, Liu
Department of Computer-Intergrated Manufacturing
Taiwan Semi-condoctor Manufacturing Co., Ltd
Hinschu, Taiwan
stdm11528@gmail.com

Shin-Jye, Lee[*]
Institute of Management of Technology
National Yang Ming Chiao Tung University
Hinschu, Taiwan
camhero@gmail.com

Xiaojun, Zeng
Department of Computer Science
The University of Manchester
Manchester, UK
x.zeng@manchester.ac.uk

*Abstract*— **In deep learning, optimization plays a vital role. By focusing on image classification, this work investigates the pros and cons of the widely used optimizers, and proposes a new optimizer: Perturbated Unit Gradient Descent (PUGD) algorithm with extending normalized gradient operation in tensor within perturbation to update in unit space. Via a set of experiments and analyses, we show that PUGD is locally bounded updating, which means the updating from time to time is controlled. On the other hand, PUGD can push models to a flat minimum, where the error remains approximately constant, not only because of the nature of avoiding stationary points in gradient normalization but also by scanning sharpness in the unit ball. From a series of rigorous experiments, PUGD helps models to gain a state-of-the-art Top-1 accuracy in Tiny ImageNet and competitive performances in CIFAR- {10, 100}. We open-source our code at link: https://github.com/hanktseng131415go/PUGD.**

## I. Introduction

An optimizer plays an essential role in updating weights to find optimum in machine learning. Since the introduction of Stochastic Gradient Descent (SGD) [1], updating in mini-batch has become the mainstream. Thus, applying SGD and other mini-batch optimizers has turned into a natural choice. Later, the proposition of SGD-based Adaptive Gradient Algorithm (Adagrad) [2] has opened another path of using the adaptive gradient to get a suitable learning rate with the individual accumulating gradients as a divider. By combining the mini-batch upgrading with the adaptive mechanism, models quickly reach a minimum. On the other hand, Normalized Gradient Descent (NGD) [3] applies $L_2$-norm of overall or individual gradient as a scaler to modify the updating step toward vector space. By the statement from NGD [3], this gradient scaling operation brings a range of properties: 1. Prevent models from stocking in saddle points and a local minimum in a loss surface (although these terrains are unlikely to exist in deep learning [4]); 2. Ameliorate the slow crawling problem of standard gradient descent near the minimum; 3. Push model to flat regions. Despite these merits, this normalization has not been seen in deep learning updating toward tensor. Thus, it inspires this work, and motivates introducing the normalization in deep learning gradient updating.

By applying gradient normalization, it is expected that having a normalized gradient for updating in each step could bring a better generalization as the discussions in Path-SGD [5] and [6]. Furthermore, each step length is bounded or controlled with a normalized gradient each time, which could further alleviate the potential of reaching a sharp minimum difficulty or sub-optimum due to stochastic [7]. In order to have bounded updating length between time to time, applying the gradient normalization in $L_2$-norm is doable when facing a 1-D or 2-D vector. However, as deep learning works on tensors (vector with inner vectors), we need to extend the normalization operator from vector to tensor. Because it is unlikely to require each component (inner vector) of tensor to be the same shape, scaling the magnitude of the gradient by $L_2$-norm is infeasible. Therefore, the second motivation of this work is developing the tensor version of NGD for deep learning.

In deep learning, while investigating generalization bound is still one of the mainstreams, sharpness-aware generalization has boomed. A series of works [8-12] has tried to explain the generalization mechanism via theoretical perspectives and proofs. By the findings of the positive relationship between sharpness-awareness and generalization [6], the same authors proposed Sharpness-Awareness Minimization (SAM) [13], which considers sharpness-aware generalization bounds and obtains state-of-the-art performances in image classifications. Based on another widely-accepted relationship between generalization and flat minimum [8, 11, 14-16], SAM and Adversarial Model Perturbation (AMP) [17] has ramped up attention toward this by showing models can reach flat minimum with the sharpness-awareness mechanisms during updating.

Summing up the above overview and extending the investigations in [6, 15, 17-21], this work focuses on inheriting the merits of gradient normalization by extending to a tensor version in deep learning and utilizing the two criteria to form a good optimizer: locally bounded updating and sharpness-awareness. Therefore, we introduce a new mini-batch gradient descent with perturbation in unit space, called

---

[*] Corresponding author

the perturbated unit gradient descent (PUGD). In brief, the main contributions in this work include:

- Usher tensor normalization for gradient updating.
- Reveal that updating within unit gradient space is locally bounded updating.
- Propose the Perturbated Unit Gradient Descent (PUGD), a gradient descent optimizer with perturbation within a unit ball. We furthery present that PUGD outperforms the existing optimizer in Fig. 1 and alleviates the chaos problem in NGDs in Fig. 2
- Carry out a series of comparisons to show that PUGD can help models obtain a state-of-the-art[1] performance in Tiny ImageNet, competitive results in CIFAR-10 and CIFAR-100 with outperforming current sharpness-aware optimizers.

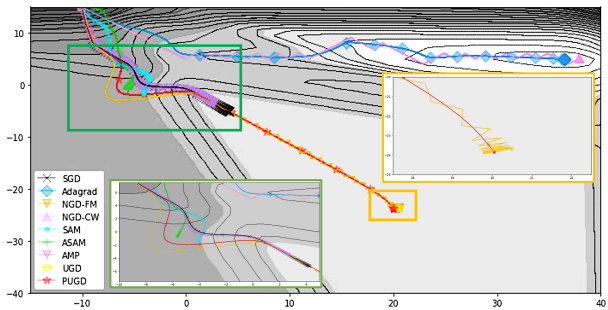

Fig. 1. Illustration of (P)UGD and other optimizers updating trajectories in MNIST 2-D landscape with mean square error. The whole picture is in the background, and the zoom-in pictures are in green and yellow rectangles. A sharp minimum is on the top right side. A flat minimum is on the bottom right side. Lastly, an area of the saddle is between the sharp and flat minimum. Every dot in the same mark represents the location of the model at t (iteration position). PUGD outperforms others with the fastest convergence, most stable trajectory, and a better sharpness-awareness. Please see detail in IV.A.4).

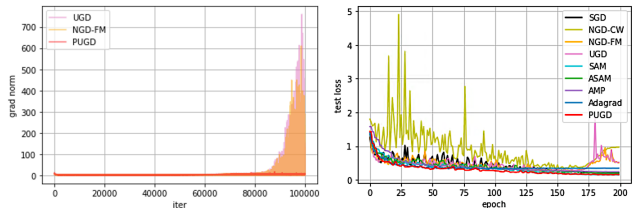

Fig. 2. UPANet16 training histories of (P)UGD and other optimizers in CIFAR-10. The histories of grad norm[2] and test loss are on the left and right, respectively. Although NGD-FM and UGD converged stably before the last iterations, dividing gradients with zero norm gradients makes chaos. PUGD, otherwise, converged stably with the lowest testing loss. Please see the same test loss results and experiment settings in IV.A.1)

The rest of this paper is divided into five parts. First, the background, motivation and a summary of PUGD have been present in this. Then, in section II, similar optimizers are listed. After, we present the mathematical proofs of (P)UGD properties among deep learning in III. Finally, a series of simulations on PUGD and competitors is shown in IV, with the conclusion in V and supplements in Appendix.

## II. RELATED WORKS

Since the introduction of SGD, the natural step of the mainstream has been choosing SGD as a default optimizer.

However, SGD has been argued with the downsides of often being unstable or reaching a bad optimum easily due to stochasticity [7]. In terms of ushering dynamic gradients, Adagrad cannot be ignored as it brought the adaptive mechanism in optimization, but the adaptive mechanism from accumulating gradients will speed up too much in the early epochs and make the gradients in later epochs vanish into zero. That is gradient vanishing by the escalating divider, so the unstable path updating becomes severe. While there are RMSProp [22] and ADAM [23], the rapid convergence speed has been negatively correlated to generalization [6]. Toward the same direction of avoiding the stationary point, NGD [3] formed its methods by applying full-magnitude (NGD-FM) and component-wise gradient norm (NGD-CW) in vector. NGD-FM, nonetheless, will easily stick the model with an extremely small divided gradient when having a great amount of gradient [3], and normalizing gradients will make convergence uneasy. On the other hand, to avoid the sticking issue in NGD-FW, NGD-CW scales every gradient component by $L_2$-norm of each component. The component gradient normalization makes every weight only update a length of one. In other words, the update length of NGD-CW is decided by the learning rate. In our experiments, component-wise gradient normalizations intend to mess the updating and end up with sub-optimum generalization performance, which is also witnessed in Adagrad, despite achieving fast-updating in early steps. Therefore, this work tries to usher the full-magnitude normalization into deep learning by formally defining the gradient normalization toward tensor, so it could also bring a better generalization.

Regarding what factor could link to better generalization, gradient flow [21] offers a mathematical perspective to prove that SGD with locally Lipschitz gradients in deep learning benefits the stable converge. With the observation in Path-SGD [5] and [6], creating a controllable updating process is positive related to generalization. Path-norm designs an updating policy based on the stable direction, balancing the difference of parameters by the norm of overall weights. Although the results with Path-norm fall behind SGD in [5], it opened a path for caring for the stability (invariant) of updating weights. Although the result of Path-SGD in [5] did not outshine SGD, the survey in [6] does support the positive relation. This work takes the same notion but creates a bounded updating process by considering overall gradient with normalization in SGD-based optimization.

Taking SGD as a based optimizer for finding a flat minimum in deep learning, the attention on local sharpness is ramping up. What has led to the surge could be SAM and AMP, which argues that the flat minimum can bring a more robust result compared with other directions, with the author's previous survey work [6] and the works [8, 11, 14-16] from the antecedents. The authors of SAM furtherly prove that it follows the PAC Bayesian Generalization bounds and can create a smooth space in empirical tests. Then, an adaptive modification of SAM is proposed, ASAM [24]. An invertible scaling operator in ASAM can help loss surface more fit to real contour. Following the same direction, Entropy-SGD [18] focuses on the local entropy of empirical loss in 2017. It utilizes the local geometry of the energy landscape to help models obtain a flat plateau with the baby steps roaming in Gibbs distribution. Some related works [25-27] investigated finding generalization well minima using local entropy. SWA [28, 29] performs the average weights traversed based on SGD in a scheduled training plan using an intuitive method to find a flat area. Furthermore, an extension of SWA [30] also obtains the efficient PAC Bayesian criteria. By building a unit sharpness-aware in UGD, the PUGD visualization of landscape did help the models to find a flat minimum. Also, the loss history shows a better generalization bound, which implies less possibility of facing an overfitting problem.

---

[1] Checked on June 29th, 2021

[2] The gradient dual-norm in $p = 2$ is the same as $L_2$-norm toward flattening every element in tensor into vector.

III. PERTURBATED GRADIENTS UPDATING WITHIN UNIT SPACE

This section proposes PUGD along with unit gradient descent (UGD, a simplified variant of PUGD) and proves its properties. As the discussion toward vector and tensor in section I, one major point of this work is to build a connection between vector and tensor by precisely defining the gradient normalization in tensor. In order to connect the existing knowledge for knowing the properties toward tensor, the proposed normalization in (P)UGD equipped with dual-norm in $p = 2$ is chosen. Therefore, a descending operation in unit space by normalizing gradient in deep learning is UGD. PUGD is proposed with a unit space perturbation to serve the purpose of sharpness-awareness in unit space. The following contents state the preliminaries and properties of (P)UGD to be a decent optimizer.

A. Notation and Preliminary

1) Notation

Throughout the work, given learning rate $\eta$ and a training set $S \triangleq \bigcup_{n=1}^{N} \{(x_n, y_n) | x_n \in \mathcal{X}, y_n \in \mathcal{Y}\} \overset{i.i.d.}{\sim} \mathcal{D}$, we let a single data-generation model $F: \mathcal{X} \to \mathcal{Y}$ parametrized from a weight vector $w \in W \subseteq \mathbb{R}^m$. Then, the loss function $\mathcal{L}_S(w) \triangleq \mathbb{E}[l(w, x_i, y_i)]$, where $i$ indicates the $i^{th}$ batch data and $l(w, x_i, y_i)$ represents the $i^{th}$ loss term under parameter $w$.

2) Preliminary

The definitions and preliminary concepts toward this work are listed and discussed firstly and below.

**Definition 1 (Dual-norm).** *If $\|\cdot\|_p$ is the $L_p$-norm on $\mathbb{R}^{M \times I \times J}$, then its dual-norm is $\|\cdot\|_q$, where $\frac{1}{p} + \frac{1}{q} = 1$. We denote $(\|U\|_p^p)^{1/q}$ as the dual-norm in (1):*

$$(\|U\|_p^p)^{\frac{1}{q}} = \sum_{m=1}^{M} \left[ \left[ \left[ \sum_{j=1}^{J_{(m)}} \sum_{i=1}^{I_{(m)}} \begin{bmatrix} u_{i_{(m)},j_{(m)}}^p & \cdots & u_{i_{(m)},J_{(m)}}^p \\ \vdots & \ddots & \vdots \\ u_{I_{(m)},j_{(m)}}^p & \cdots & u_{I_{(m)},J_{(m)}}^p \end{bmatrix} \right]^{\frac{1}{p}} \right]^q \right]^{\frac{1}{q}}$$

$$= \sup_{\|E\| \leq 1} U^T V, \text{ for all } U, V \in \mathbb{R}^{M \times I \times J}.$$

(1)

*where $U_m \in \mathbb{R}^{I_{(m)} \times J_{(m)}}$ is composed of $u_{i_{(m)},j_{(m)}}^p$, $u_{i_{(m)},j_{(m)}}^p$ indicates an element among $m^{th}$ matrix of a matrix with row index $i_{(m)}$, column index $j_{(m)}$, and is powered by $p$. When listing each flattened $U_m$ in a matrix, the dual-norm forms the $L_P$-norm in $\|U\|_p$ as (2):*

$$\|U\|_p = \left[ \sum_{m=1}^{M} U_m \right]^{1/q} \quad (2)$$

$$= \left[ \sum_{m=1}^{M} \sum_{j=1}^{J} \sum_{i=1}^{I} \left[ u_{i_{(m)},j_{(m)}}^p \cdots u_{I_{(M)},J_{(M)}}^p \right] \right]^{1/q}.$$

*When $p = q = 2$, we simply denote the dual-norm as $\|\cdot\|$ along with the $L_2$-norm in $\|\cdot\|_2$.*

When having $L_2$-norm in $\|\cdot\|_2$, dual-norm in $\|\cdot\|$, and a tensor $U = [[1, 2, 3], [4, 5]]$, we say that $X$ is a group of vectors with various component shapes in vector, and $u_1 = [1, 2, 3]$ and $u_2 = [4, 5]$ are the components in vector. In this case, calculating components in $L_2$-norm becomes infeasible, since it is facing a vector instead of a single value among each component. Namely, directly getting $L_2$-norm of $U$ is impropriate. On the other hand, because a dual of dual-norm is the original norm [31](Theorem 13.1), calculating dual-norm in $L_2$ of $U$ turns into $\|U\| = \sqrt{\sqrt{u_1^2}^2 + \sqrt{u_2^2}^2} = \sqrt{\|u_1\|_2^2 + \|u_2\|_2^2}$. This operator makes norm of element-wise in tensor possible. It should be noticed that, for a tensor, the (dual-) norm when $p = 2$ and $q = 2$ defined in the above is the square root of the simple square sum of each element in the tensor. This is not true in general when $p$ and $q$ are not equal to 2.

**Definition 2 (Unit tensor).** *Let $v \in \mathcal{V}$ be a tensor in tensor space $\mathcal{V}$, and $\|\cdot\|$ be the dual-norm of the tensor, then the tensor $\hat{v} = \frac{v}{\|v\|}$ is the unit tensor of $v$, and $\|\hat{v}\| = 1$.*

SGD has brought online learning with stochastic mini-batch learning into metadata learning to set a proper learning rate. The typical updating process with stochastically mini-batch updating can be presented in (3) with the right $\eta_t$ to reach the optimum. $\eta_t$ is referred to the size of updated steps at different times, $t$. The descent process follows:

$$w_{t+1} = w_t - \eta_t g_t. \quad (3)$$

where $w_t, \eta_t,$ and $g_t$ denotes the weight, learning rate, and gradients from $\mathcal{L}_S$, respectively.

Based on the same notion in NGD [3] and discussions [5, 6], having a bounded updating throughout the training process is essential. Under the same perspective, applying full-magnitude gradient normalization with $L_2$-norm as NGD can help gain that. Although there is no formal work to define the algorithm, NGD (in the full-magnitude gradient) can be represented as:

$$w_{t+1} = w_t - \eta_t \frac{g_t}{\|g_t\|_2}, \quad (4)$$

where $\frac{g_t}{\|g_t\|_2}$ is the normalizing gradients with $L_2$-norm. By this gradient normalization, the updating tendency is expected to address the stochastic behaviour of SGD and prevent to stuck in stationary points. However, although the advantages of having a stable gradient difference and avoiding local minimum, this work shows that normalizing gradients could affect the final convergence, which might dim the stable updating in the final iterations. From another standpoint, Path-SGD argues that SGD tends to update networks poorly with unbalanced updated weights. The path-norm in Path-SGD can be seen as a weight normalization and is argued to be positive with generalization in a survey [6]. Therefore, this work picks up a similar idea toward tensor to introduce a gradient normalization in dual-norm as the unit gradients and tries to explain its benefits toward the stochastic issue. We argue that unit gradients are also a regularization in the gradient perspective. It can bound the difference of updating and thus alleviate the issue of unstable updating due to stochasticity.

In the light of sharpness-awareness, (A)SAM and AMP have attracted much attention to date. Different from the typical method of minimizing one step empirical risk loss, sharpness-aware optimizers minimize the following PAC-Bayesian generalization bound based on the sharp risk $\epsilon^*$ from the second step with the following expression (i.e., SAM):

$$\mathcal{L}_\mathcal{D}(w) \leq \max_{\|\epsilon\|\leq\rho} \mathcal{L}_\mathcal{S}(w+\epsilon) + h(\frac{\|w\|_2^2}{\rho^2}), \tag{5}$$

where $h: \mathbb{R}^+ \to \mathbb{R}^+$ is a strictly increasing function. Therefore, the goal is to minimize a dual-norm ball with $\rho$, which is another hyperparameter to decide the size of sharpness-aware space, and then this minimization can further be solved by finding the max perturbation $\epsilon^*$ of $\mathcal{L}_\mathcal{S}(w+\epsilon)$:

$$\begin{aligned}\epsilon^* &= arg\max_{\|\epsilon\|\leq\rho} \mathcal{L}_\mathcal{S}(w+\epsilon) \\ &\approx \rho \frac{\nabla\mathcal{L}_\mathcal{S}(w)}{\|\nabla\mathcal{L}_\mathcal{S}(w)\|}.\end{aligned} \tag{6}$$

Optimizers can help models choose a flat direction by observing sharpness geometry around a radius. Nonetheless, setting the radius or the sharpness distribution requires prior knowledge, which implies a careless setting could easily pollute the performance. Different from the mentioned perturbated optimizers, we comment that a perturbation with a similar adaptive mechanism with ASAM's as (7),

$$\max_{\|T_w^{-1}\epsilon\|_p\leq\rho} \mathcal{L}_\mathcal{S}(w+\epsilon) - \mathcal{L}_\mathcal{S}(w) \tag{7}$$

where $T_w^{-1}$ is the normalization operator which does not change the loss function, in a unit gradient space can generalize well.

This work answers the needed criteria as a decent optimizer: locally bounded updating and sharpness-awareness, in the following parts.

### B. Unit Gradient Descent

Applying the unit vector into typical gradient descent (3) makes sure that the gradient norm becomes one based on Definition 2. However, there is no formal optimization of applying normalizing gradients in deep learning toward tensor. As discussed in III.A.2) and Definition 1, $L_2$-norm is a special case in dual-norm. Following, we formally define the tensor gradient normalization in descending as Definition 3.

**Definition 3 (Unit gradient descent, UGD).** *Let $g_t = \nabla\mathcal{L}(w_t)$ be the gradients of the loss function at $t$, and then be divided with the overall norm of gradient. We form a gradient descent in unit gradient space:*

$$\begin{aligned}w_{t+1} &= w_t - \eta_t \frac{g_t}{\|g_t\|} \\ &= w_t - \eta_t \widehat{g_t},\end{aligned} \tag{8}$$

*where $\|g_t\|$ is dual-norm of full magnitude of gradients, $\frac{g_t}{\|g_t\|}$ is the unit gradients corresponding to each component, and $\|\widehat{g_t}\| = 1$.*

*1) Difference between preliminaries*

In (8), the formal equation of gradient normalization in tensor is defined. Compared with (4), the conventional gradient normalization in machine learning is using (2), which is a special case of (1) and has a drawback of being unable to deal with tensor. The different shape makes operating impossible, as $L_2$-norm calculates the root of the sum from each square element. Nonetheless, each square element in the tensor is not a single value but a matrix. Applying $\|\cdot\|$, it makes this operation possible.

On the side of using dual-norm in deep learning, we have seen (A)SAM and AMP that also use dual-norm as a high-dimensional space scaler for sharpness detecting, $\|\nabla\mathcal{L}_\mathcal{S}(w)\|$ in (6). However, this operation seems to generate landscape noise instead of updating the scaler. The above moves inspire this work, and then we propose a bounded updating with adaptive operation and perturbation both within unit space by applying dual-norm within gradients. The following content only focuses on and discusses tensors in deep learning.

*2) Locally bounded updating*

To achieve a bounded difference during updating, having a bounded difference with following uniform distribution as Definition 4 is expected.

**Definition 4 (Uniform bounded difference).** *Let $d_t$ be a sequence function of the difference between two steps such that:*

$$d_t = \|f_t - f_{t+1}\|. \tag{9}$$

*If there exists a constant $C \geq 0$ for all $t$:*

$$\|d_t\| \leq C, \tag{10}$$

*we say that $d_t$ is uniformly bounded difference.*

Different from Path-SGD, we expect to obtain the stability by caring the locally difference of gradients among steps $t \to t+1$, as Definition 4. The benefits of caring locally gradients toward convergence have also been argued in [21]. In that case, we present that gradients among $w_t$ and $w_{t+1}$ following the uniform bounded difference under UGD form a locally bounded updating as Theorem 1 below.

**Theorem 1 (Locally bounded updating).** *Given $\nabla\mathcal{L}(w)$, the loss function under UGD constructs a locally stable difference of gradients for any $w$ at step $t \to t+1$:*

$$d_t = \left\| \frac{\nabla \mathcal{L}(w_t)}{\|\nabla \mathcal{L}(w_t)\|} - \frac{\nabla \mathcal{L}(w_{t+1})}{\|\nabla \mathcal{L}(w_{t+1})\|} \right\| \leq C. \tag{11}$$

Please see the proof in Appendix. A

The proof indicates locally bounded updating in a gradient unit space. The identical phenomenon has been observed in Fig. 1 and Fig. 2, where UGD stably converge (before the chaotic in the final iterations).

### C. Perturbated Unit Gradient Descent

Although the locally bounded updating within unit gradients has been proved in Theorem 1, the chaotic issue in the final convergence steps remains, see Fig. 1 and Fig. 2. A typical suggestion for this dilemma is applying early-stop (also called elbow method) to prevent. Unfortunately, such a scheme only cures the symptoms, not the disease. Toward this disease, we offer a solution by extending the basics of UGD with the intuitive solution from the practical perspective: perturbation. More precisely, we propose unit perturbation and merge it into UGD to become Perturbated Unit Gradient Descent (PUGD). By strictly obeying Definition 2, the unit perturbation is designed with an adaptive mechanism as ASAM but scans in a unit radius and updates the gradients in a unit ball at any $t$. Furthermore, the unit perturbation keeps the dual-norm of gradients non-zero, so it will prevent gradients bouncing up because of getting zero gradients as the divider. Eventually, we define PUGD as Definition 4.

**Definition 4 (Perturbated Unit gradient descent, PUGD).** *Let $g_t = \nabla \mathcal{L}(w_t)$ be the gradients of the loss function at $t$ with perturbation and descending within unit space:*

$$\begin{cases} \hat{\epsilon}_t = \frac{|w_t| \cdot g_t}{\||w_t| \cdot g_t\|} \\ g_{t^*} = \nabla f(w_t + \hat{\epsilon}_t) \\ w_{t+1} = w_t - \eta_t \frac{(g_{t^*} + g_t)}{\|g_{t^*} + g_t\|} = w_{c,t} - \eta_t U_t \end{cases}, \tag{12}$$

where $g_{t^*}$ is the gradients from the unit perturbation $\epsilon_t$ with adaptive steps toward each component in a unit ball within the norm of total perturbation radius $\|\hat{\epsilon}_t\| = 1$, $U_t = \frac{(g_{t^*} + g_t)}{\|g_{t^*} + g_t\|}$ is the unit gradient at $t$, and the step length of unit gradients $\|U_t\| = 1$. As a result, the adaptive sharpness-aware process within a unit space can be further expressed as Theorem 2 to become unit sharpness-aware optimizer, analogue [13] (Theorem 1).

**Theorem 2 (Unit sharpness-awareness, state informally).** *For any distribution $\mathcal{D}$, with probability $1 - \delta$ over the choice of the training set $\mathcal{S} \sim \mathcal{D}$,*

$$\mathcal{L}_\mathcal{D}(w) \leq \max_{\|\epsilon\|=1} \mathcal{L}_\mathcal{S}(w + \epsilon) + h\left(\frac{\|w\|_2^2}{\eta^2}\right), \tag{13}$$

*where $h: \mathbb{R}^+ \to \mathbb{R}^+$ is a strictly increasing function (under some technical conditions on $\mathcal{L}_\mathcal{D}(w)$).*

Please see the detail proof in Appendix. B

Finally, the whole algorithm is demonstrated in Fig. 3 and the following pseudo-code. In PUGD, Unit sharpness-awareness does not require setting a fitting radius as SAM and an inner learning rate as AMP. Also, the final converge path (red arrow in Fig. 3) has thoroughly considered the surrounding geometry, inheriting the merits of UGD.

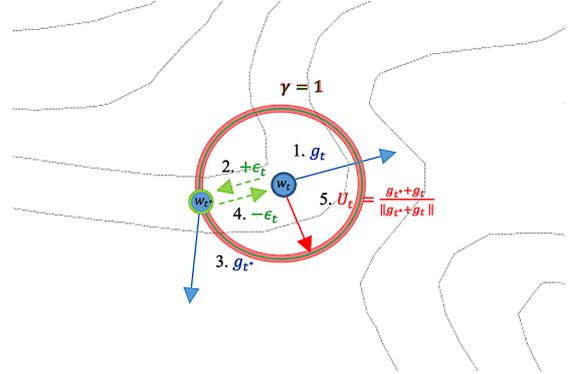

Fig. 3. Demonstration of PUGD ignoring learning rate η.

---

**Algorithm** Perturbated Unit Gradient Descent (PUGD)

**Input:** Training set $\mathcal{S} \triangleq \cup_{n=1}^N \{(x_n, y_n)\}$ with the loss function $\mathcal{L}_\mathcal{S}(w) \triangleq \mathbb{E}[l(w, x_i, y_i)]$, where $i$ indicating $i^{th}$ batch data and $l(w, x_i, y_i)$ represents $i^{th}$ loss term under parameter $w$ by setting random initial weight $w_0$ with any $\eta_t \in (0,1]$, given $t = 0$.
**Output:** Trained weight $w_t$ with PUGD
**while** *not converged* **do:**
    For $i$ in random sample batch $B: \{(x_1, y_1), \cdots, (x_n, y_n)\}$:
1. Compute the gradients from a batch set $s$:
$$g_t = \nabla_w \mathcal{L}_s(w_t);$$
2. Noise the weight by manually adding perturbation:
$$w_{t^*} = w_t + \hat{\epsilon}_t,$$
where $\hat{\epsilon}_t = \frac{|w_t| \cdot g_t}{\||w_t| \cdot g_t\|}$;
3. Compute the perturbation gradients of the same set $s$:
$$g_{t^*} = \nabla \mathcal{L}_s(w_t + \hat{\epsilon}_t);$$
4. Call back to $w_t$:
$$w_t = w_{t^*} - \hat{\epsilon}_t$$
5. Descend by using the unit gradients:
$$w_{t+1} = w_t - \eta_t U_t,$$
where $U_t = \frac{g_{t^*} + g_t}{\|g_{t^*} + g_t\|}$;

    $t = t + 1$
**end while**
**return** $w_t$

---

## IV. EXPERIMENTS

In this section, we implemented three experiments, including (1) optimizing behaviours comparison, (2) end-to-end comparison, and (3) finetuning comparison. The experiments were implemented on open datasets in the equipment with RTX-Titan, 32G RAM, and an eight-core processor. In addition, every experiment followed the hyperparameters setting in the learning rate (LR) = 0.1, weight decay = 0.0005, momentum = 0.9 (if there is), Nesterov = False, and the learning schedule of cosine annealing [32]. Every subject followed the same Pytorch default initial weights policy, a random uniform distribution (-weight number, weight number).

### A. Optimizing behaviors

To observe the behaviours of (P)UGD and others, this subsection of simulation has been organized into four-folds: IV.A.1) CIFAR-10 training history; IV.A.2) CIFAR-10 loss landscapes in 3D; IV.A.3) CIFAR-10 updating trajectories in 2-D; IV.A.4) MNIST updating trajectories with MLPs in 2-D;

by highly referring to the visualization loss landscape work [33] and animation code in the link [3]. Through these observations, we can get a picture of what (P)UGD can do with models.

*1) CIFAR-10 training history*

When it comes to understanding the training behaviour, analyzing the training history in 1-D is one of the widely-used and easy options. Therefore, we presented the loss and accuracy history together from SGD, Adagrad, NGDs, (A)SAM, and (P)UGD in Fig. 4 to Fig. 5. Among the loss plots, the gap represents the difference between training loss and testing loss, so it could also indicate the overfitting problem when it surged. We implemented all subjects by following the experiment settings at the beginning of IV. Every performance was recorded from 100 epochs in perturbated optimizers: (A)SAM and PUGD, 200 epochs in non-perturbated ones: SGD, Adagrad, NGDs, and UGD. Also, every subject followed the same batch size in 100 by training UPANet16 [34] for having faster speed and updating smooth path every time. Finally, the setting from each optimizer followed the default and recommended parameters from the proposed authors.

normalization. Later, the result from NGD-FM and UGD share the same pattern, a hybrid with smooth histories before the $175^{th}$ epoch and vibrate histories after that. At the end of the spectrum, the smooth history of Adagrad is a sugar coat that has a high loss. In the rest of the outcomes from (A)SAM to PUGD, (A)SAM are promising with smooth history and low loss. However, they all seem not to have the ability to bound the testing loss under the training loss as PUGD did. Most importantly, the gap of PUGD decreased in the last epochs. To summarize this part, PUGD is much better than others from the training histories.

*2) CIFAR-10 loss landscapes in 3D*

Obeying the same setting as the last part, we extended the 1-D loss history to the 3-D space to show a transparent optimizing environment where each updating method was created in CIFAR-10 for the model. Please see Fig. 6 to Fig. 7.

The landscape from SGD guarantees a smooth landscape in UPANet16, but Adagrad turned the landscape to a steep view. Moreover, NGD-CW made it full of valleys, so we can deduce that the component-wise method makes the space fluctuate. Conversely, smoother ones can be seen from (A)SAM, which is consistent with the experimental result in SAM [13]. As the last part observation, PUGD shows a smoother one than (A)SAM and ASAM's. Moreover, the landscape from PUGD has the lowest overall loss. From this observation, PUGD can create a smoother updating area for models from viewing loss landscape.

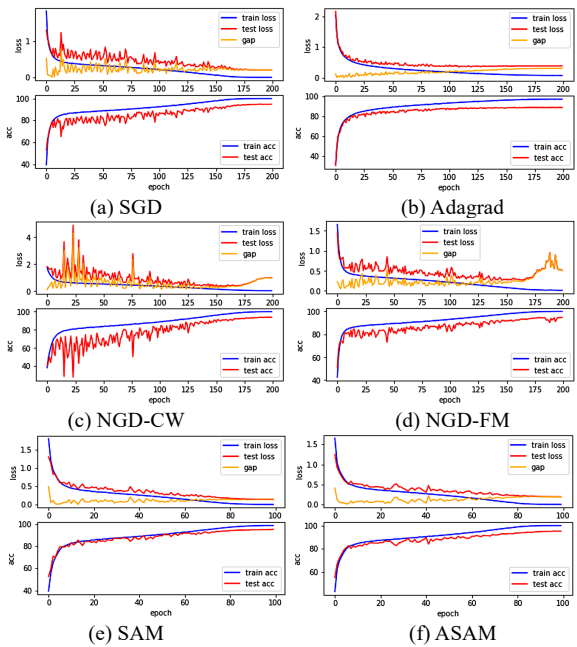

Fig. 4. Training histories of other optimizers.

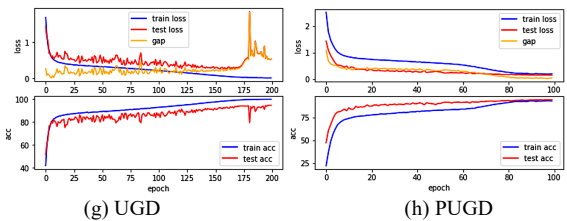

Fig. 5. Training histories of (P)UGD.

The training histories can be split into two types of patterns, smooth and fluctuated, based on taking SGD as the standard. Among the fluctuating training histories, SGD is in line with the argument of the stochasticity issue. Except for SGD, NGD-CW experienced a severe fluctuated history, which might be due to the amplification from component-wise

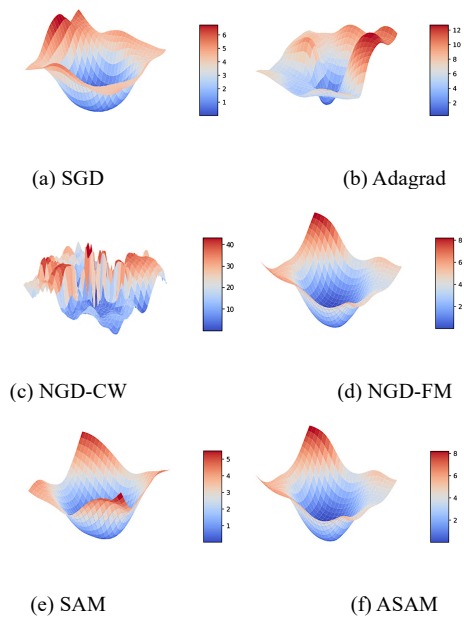

Fig. 6. Visualization loss landscapes of other optimizers.

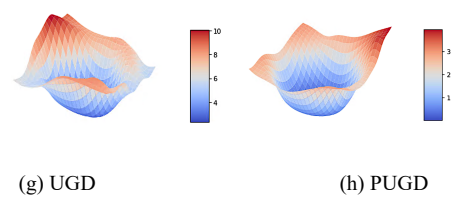

Fig. 7. Visualization loss landscapes of (P)UGD.

*3) CIFAR-10 updating trajectories in 2-D*

Combing a 1-D loss history with a 3-D landscape and then projecting it to the 2-D landscape, we can obtain the possible updating trajectories based on different optimizers. We present trajectories with the same setting as the last two parts but in a 2-D space with a trajectory log in each plot. Please see Fig. 8 and Fig. 9.

In these contours, we can furtherly assure that component-wise adaptive gradient operation will aggregate the fluctuation of the space and, possibly, turn the flat landscape into a steep one, from Adagrad and NGD-CW results. Apart from that, PUGD has the flattest convergence area as the last two observations. Therefore, we can answer that PUGD created a great generalization bound and flat updating environment. However, we cannot get more information about each optimizer lead a model to a minimum from sharing a similar path, so a simulation of gathering subjects in the same landscape was conducted in IV.A.4).

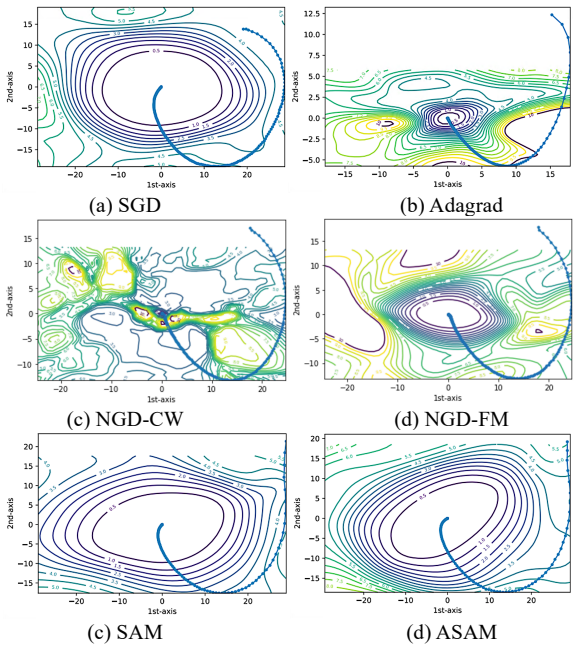

Fig. 8. The contours and trajectories of other optimizers.

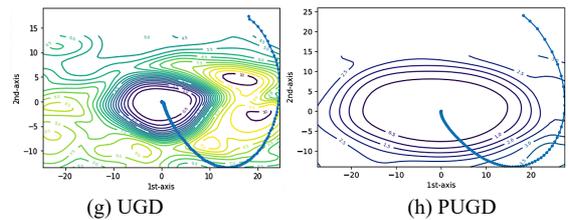

Fig. 9. The contours and trajectories of (P)UGD.

*4) MNIST updating trajectories with MLPs in 2D*

To gain a better picture, we gathered all competitors in the same fixed loss landscape. Different from the referenced code[4], we plotted the terrain of the first 100$^{th}$ training data (the green-blue landscape) and the terrain of the first 100$^{th}$ testing data (the blue-red landscape) from MNIST together. In this way, we can compare the reaching training area with the testing one. When the time spot of the model is located in a blue area of training but has a red area of testing, it indicates the optimizer risk a model in an overfitting situation. An ideal direction is a way to a plateau (flat minimum) [15, 18] to have a robust result and avoid potential overfitting problems. Because of involving eight optimizers, every optimizer was iterated 10000 times in total, except perturbated ones {(A)SAM, AMP, PUGD} in 5000, by sampling every 100 iterations from the same start point in [-10.1, -15], which can be viewed as the initial weight of the model. The rest of the experiment settings remained identical to the referenced code. We presented the 3D trajectories in Fig. 10 and the 2D ones in Fig. 1.

The patterns of this simulation can be roughly divided into three types based on their final locations: the sharp minima, flat minimum, and halfway. This simulation expects a decent optimizer to lead the model to the flat minimum where the green area is in training and the blue area is in testing.

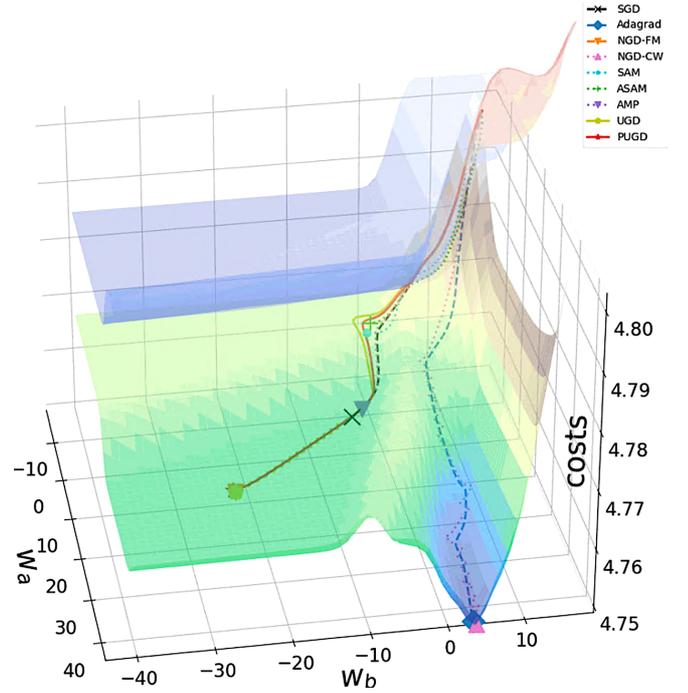

Fig. 10. The 3D trajectories in MNIST.

Among the endpoints on the sub-optimum, it contains most of the existing optimizers. In the sharp minimum with potential overfitting problem in testing, the end location of Adagrad is in line with the above discussions. However, it is surprising that NGD-CW reached the exact location. The underlying reason for having a sharp minimum in Adagrad and NGD-CW could be the fast descending and vanishing gradients of component-wise gradient normalization. On the other side, SGD is halfway to the flat area. While existing perturbation- or sharpness-awareness-based optimizers claimed superior mechanisms in the leading model, we did not see a clear picture of superiority in this simulation. These sharpness-aware optimizers avoided a sharp region where SGD stepped in, but they ended up marching on a highland with higher training and testing loss. Lastly, it is also surprising that AMP neither outshined SGD nor sheerness-aware in this experiment.

Although (P)UGD and NGD-FM share similar trajectories, PUGD has a stabler updating and surpasses others. Firstly, UGD shared the same trajectory as the NGD-FM, which is deducible because of equality between dual-norm and $L_2$-norm when $p = 2$. At the same time, NGD-FM and UGD suffer difficult convergence issues, similar to Fig. 2. We answered this issue with the unit perturbation in PUGD that fixed the issue and helped to sharpness-aware as the red trajectory in the yellow rectangle in Fig. 1. From this simulation, we can furthermore assure that PUGD pushed the model to the flat minimum stably, referencing the green rectangle in Fig. 1. Moreover, this also completed why PUGD can create a smooth landscape and lead to a broader flat minimum in the above simulations.

*B. Evaluation in end-to-end training*

In this sub-section, we conducted PUGD in CIFAR-10 and CIFAR-100 compared with the optimizers in IV.A.4). The experimental setting followed the description in IV.A.1). Meanwhile, a series of models have been used to evaluate the optimizers effect toward various structures, including the types of deep like VGGs, a residual connection like ResNets, a dense connection like DenseNets, and a hybrid connection UPANets. The recorded average and standard deviation performances were obtained from three testing results on the test data. Please see the below comparisons from TABLE IV and TABLE V in Appendix.

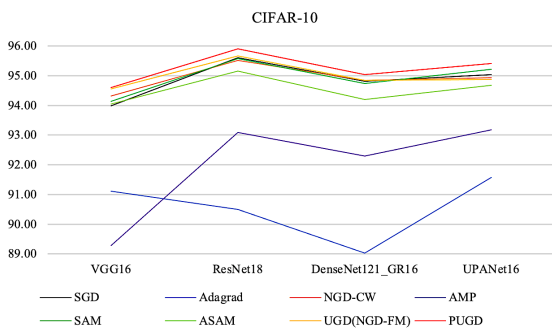

Fig. 11. CIFAR-10 end-to-end training results.

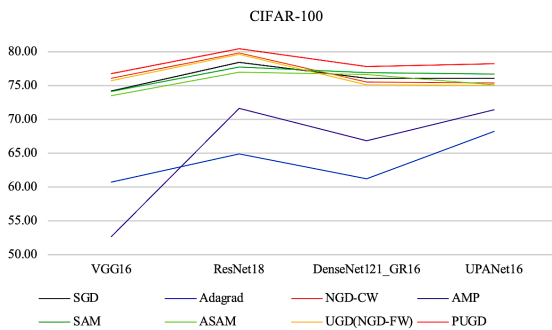

Fig. 12. CIFAR-100 end-to-end training results.

From Fig. 11 and Fig. 12, despite the difference in epochs for doubling time in perturbation-based optimizers training, it can easily be seen that the results from PUGD have improved. Moreover, PUGD has the best performance among these. The results are in line with the updating behaviour of PUGD in marching on the plateau, which assures better performance even in more complex tasks, CIFAR-100.

*C. Evaluation in finetuning*

The goal of transfer learning [35] is to improve the performance of the target learner by training a previous model on a large (or similar) dataset and then transferring the knowledge from the previous model to the target learner. Because transfer learning is a powerful technique, we used the pre-trained models (on 1K-ImageNet [36]) to finetune CIFAR-10, CIFAR-100, and Tiny ImageNet by the proposed method PUGD. The most experimental setting followed the setting in IV.B except the hyperparameters: learning rate (LR) = $\{0.01, 0.001, 0.005\}$ and input size = $(224, 224)$. Then, more models were finetuned and recorded in the best accuracy as the following.

The finetuning outcomes of PUGD in Tiny ImageNet obtained state-of-the-art performance to the best of our knowledge. Also, compared with the result from the paper of SAM, the finetuning results of PUGD outperformed SAM significantly, especially in CIFAR-100. We also learned another lesson: despite testing on varied models, improvement is guaranteed in PUGD. That reassures the superiority of PUGD in our comparisons.

TABLE I. CIFAR-10 FINETUNING RESULTS (TOP-1%)

| Dataset<br>Model (LR) | CIFAR-10 |
|---|---|
| ResNet18 (0.01) | 97.46 |
| ResNet50 (0.01) | 97.95 |
| ResNet101 (0.01) | 98.40 |
| ResNet152 (0.01) | 98.50 |
| ViT-B/16 (SAM [13]) | 98.60 |
| ViT-B/16 (0.001) | 99.13* |
| DeiT-B/16 (0.005) | 98.74 |

* The current top fifth checked on June 29th, 2021

TABLE II. CIFAR-100 FINETUNING RESULTS (TOP-1%)

| Dataset<br>Model (LR) | CIFAR-100 |
|---|---|
| ViT-B/16 (SAM [13]) | 89.1 |
| ViT-B/16 (0.001) | 93.95* |

* The current top fifth checked on June 29th, 2021

TABLE III. TINY IMAGENET FINETUNING RESULTS (TOP-1%)

| Dataset<br>Model (LR) | Tiny ImageNet |
|---|---|
| ViT-B/16 (0.001) | 90.74* |
| DeiT-B/16 (0.005) | 91.02* |

* The SOTA checked on until June 29th, 2021.

V. CONCLUSION

This work proposed a perturbated gradient updating within unit space for deep learning focusing on image classification. First, a bridge between vector and tensor is built by formally introducing the gradient normalization in tensor with dual-norm. Later, the proofs explained locally bounded updating gradients in unit space and sharpness-awareness of PUGD. The implemented simulations also show that PUGD can reach a flat minimum and outperform others with SOTA performances. We hope this work can raise the attention of the communities to research the unit tensor in deep learning given having better performance and bounded updating difference during optimization.

## References

[1] L. Bottou, "Large-scale machine learning with stochastic gradient descent," in *Proceedings of COMPSTAT'2010*: Springer, 2010, pp. 177-186.

[2] J. Duchi, E. Hazan, and Y. Singer, "Adaptive subgradient methods for online learning and stochastic optimization," *Journal of machine learning research,* vol. 12, no. 7, 2011.

[3] J. Watt, R. Borhani, and A. K. Katsaggelos, *Machine learning refined: foundations, algorithms, and applications*. Cambridge University Press, 2020.

[4] K. Kawaguchi, "Deep learning without poor local minima," *arXiv preprint arXiv:1605.07110,* 2016.

[5] B. Neyshabur, R. Salakhutdinov, and N. Srebro, "Path-sgd: Path-normalized optimization in deep neural networks," *arXiv preprint arXiv:1506.02617,* 2015.

[6] Y. Jiang, B. Neyshabur, H. Mobahi, D. Krishnan, and S. Bengio, "Fantastic generalization measures and where to find them," *arXiv preprint arXiv:1912.02178,* 2019.

[7] A. Géron, *Hands-on machine learning with Scikit-Learn, Keras, and TensorFlow: Concepts, tools, and techniques to build intelligent systems*. O'Reilly Media, 2019.

[8] G. K. Dziugaite and D. M. Roy, "Computing nonvacuous generalization bounds for deep (stochastic) neural networks with many more parameters than training data," *arXiv preprint arXiv:1703.11008,* 2017.

[9] P. M. Long and H. Sedghi, "Generalization bounds for deep convolutional neural networks," *arXiv preprint arXiv:1905.12600,* 2019.

[10] V. Nagarajan and J. Z. Kolter, "Deterministic PAC-bayesian generalization bounds for deep networks via generalizing noise-resilience," *arXiv preprint arXiv:1905.13344,* 2019.

[11] B. Neyshabur, S. Bhojanapalli, D. McAllester, and N. Srebro, "Exploring generalization in deep learning," *arXiv preprint arXiv:1706.08947,* 2017.

[12] C. Wei and T. Ma, "Data-dependent sample complexity of deep neural networks via lipschitz augmentation," *arXiv preprint arXiv:1905.03684,* 2019.

[13] P. Foret, A. Kleiner, H. Mobahi, and B. Neyshabur, "Sharpness-aware minimization for efficiently improving generalization," *arXiv preprint arXiv:2010.01412,* 2020.

[14] S. Hochreiter and J. Schmidhuber, "Simplifying neural nets by discovering flat minima," in *Advances in neural information processing systems*, 1995, pp. 529-536.

[15] N. S. Keskar, D. Mudigere, J. Nocedal, M. Smelyanskiy, and P. T. P. Tang, "On large-batch training for deep learning: Generalization gap and sharp minima," *arXiv preprint arXiv:1609.04836,* 2016.

[16] L. Dinh, R. Pascanu, S. Bengio, and Y. Bengio, "Sharp minima can generalize for deep nets," in *International Conference on Machine Learning*, 2017: PMLR, pp. 1019-1028.

[17] Y. Zheng, R. Zhang, and Y. Mao, "Regularizing neural networks via adversarial model perturbation," in *Proceedings of the IEEE/CVF Conference on Computer Vision and Pattern Recognition*, 2021, pp. 8156-8165.

[18] P. Chaudhari *et al.*, "Entropy-sgd: Biasing gradient descent into wide valleys," *Journal of Statistical Mechanics: Theory and Experiment,* vol. 2019, no. 12, p. 124018, 2019.

[19] P. Izmailov, D. Podoprikhin, T. Garipov, D. Vetrov, and A. G. Wilson, "Averaging weights leads to wider optima and better generalization," *arXiv preprint arXiv:1803.05407,* 2018.

[20] B. Guedj, "A primer on PAC-Bayesian learning," *arXiv preprint arXiv:1901.05353,* 2019.

[21] Z. Ji and M. Telgarsky, "Directional convergence and alignment in deep learning," *arXiv preprint arXiv:2006.06657,* 2020.

[22] G. Hinton, N. Srivastava, and K. Swersky, "Neural networks for machine learning lecture 6a overview of mini-batch gradient descent," *Cited on,* vol. 14, no. 8, p. 2, 2012.

[23] D. P. Kingma and J. Ba, "Adam: A method for stochastic optimization," *arXiv preprint arXiv:1412.6980,* 2014.

[24] J. Kwon, J. Kim, H. Park, and I. K. Choi, "ASAM: Adaptive Sharpness-Aware Minimization for Scale-Invariant Learning of Deep Neural Networks," *arXiv preprint arXiv:2102.11600,* 2021.

[25] C. Baldassi, A. Ingrosso, C. Lucibello, L. Saglietti, and R. Zecchina, "Subdominant dense clusters allow for simple learning and high computational performance in neural networks with discrete synapses," *Physical review letters,* vol. 115, no. 12, p. 128101, 2015.

[26] P. Baldi and K. Hornik, "Neural networks and principal component analysis: Learning from examples without local minima," *Neural networks,* vol. 2, no. 1, pp. 53-58, 1989.

[27] S. Zhang, A. Choromanska, and Y. LeCun, "Deep learning with elastic averaging SGD," *arXiv preprint arXiv:1412.6651,* 2014.

[28] G. Yang, T. Zhang, P. Kirichenko, J. Bai, A. G. Wilson, and C. De Sa, "SWALP: Stochastic weight averaging in low precision training," in *International Conference on Machine Learning*, 2019: PMLR, pp. 7015-7024.

[29] B. Athiwaratkun, M. Finzi, P. Izmailov, and A. G. Wilson, "There are many consistent explanations of unlabeled data: Why you should average," *arXiv preprint arXiv:1806.05594,* 2018.

[30] W. J. Maddox, P. Izmailov, T. Garipov, D. P. Vetrov, and A. G. Wilson, "A simple baseline for bayesian uncertainty in deep learning," *Advances in Neural Information Processing Systems,* vol. 32, pp. 13153-13164, 2019.

[31] R. Tibshirani, "Duality Uses and Correspondences," 2016, pp. 13-3.

[32] I. Loshchilov and F. Hutter, "Sgdr: Stochastic gradient descent with warm restarts," *arXiv preprint arXiv:1608.03983,* 2016.

[33] H. Li, Z. Xu, G. Taylor, C. Studer, and T. Goldstein, "Visualizing the loss landscape of neural nets," *arXiv preprint arXiv:1712.09913,* 2017.

[34] C.-H. Tseng *et al.*, "UPANets: Learning from the Universal Pixel Attention Networks," *arXiv preprint arXiv:2103.08640,* 2021.

[35] F. Zhuang *et al.*, "A comprehensive survey on transfer learning," *Proceedings of the IEEE,* vol. 109, no. 1, pp. 43-76, 2020.

[36] J. Deng, W. Dong, R. Socher, L.-J. Li, K. Li, and L. Fei-Fei, "Imagenet: A large-scale hierarchical image database," in *2009 IEEE conference on computer vision and pattern recognition*, 2009: Ieee, pp. 248-255.

[37] N. S. Chatterji, B. Neyshabur, and H. Sedghi, "The intriguing role of module criticality in the generalization of deep networks," *arXiv preprint arXiv:1912.00528,* 2019.

## APPENDIX

### A. Proof of Theorem 1

Suppose $\nabla \mathcal{L}(w_t) \cdot \nabla \mathcal{L}(w_{t+1}) \neq 1$, we obtain the difference of gradients under UGD at any step $t \to t+1$:

$$d_t = \|\nabla \mathcal{L}(w_t) - \nabla \mathcal{L}(w_{t+1})\|. \quad (1)$$

Based on Definition 1, the difference of gradients under UGD can be further expressed as:

$$\begin{aligned} d_t &= \left\| \frac{\nabla \mathcal{L}(w_t)}{\|\nabla \mathcal{L}(w_t)\|} - \frac{\nabla \mathcal{L}(w_{t+1})}{\|\nabla \mathcal{L}(w_{t+1})\|} \right\| \\ &= \sqrt{\left(\frac{\nabla \mathcal{L}(w_t)}{\|\nabla \mathcal{L}(w_t)\|} - \frac{\nabla \mathcal{L}(w_{t+1})}{\|\nabla \mathcal{L}(w_{t+1})\|}\right)^2} \\ &= \sqrt{1 + 1 - 2\frac{\nabla \mathcal{L}(w_{t+1})}{\|\nabla \mathcal{L}(w_{t+1})\|} \cdot \frac{\nabla \mathcal{L}(w_t)}{\|\nabla \mathcal{L}(w_t)\|}} \\ &= \sqrt{2 \times \left(1 - \frac{\nabla \mathcal{L}(w_{t+1})}{\|\nabla \mathcal{L}(w_{t+1})\|} \cdot \frac{\nabla \mathcal{L}(w_t)}{\|\nabla \mathcal{L}(w_t)\|}\right)}. \end{aligned} \quad (2)$$

Suppose $\nabla \mathcal{L}(w_t) \cdot \nabla \mathcal{L}(w_{t+1}) \neq 1$, we obtain (3):

$$-1 \leq \frac{\nabla \mathcal{L}(w_t) \cdot \nabla \mathcal{L}(w_{t+1})}{\|\nabla \mathcal{L}(w_t)\| \|\nabla \mathcal{L}(w_{t+1})\|} < 1, \quad (3)$$

by applying *Cauchy-Schwarz inequality*, so (2), with the inequality (3), it can further be manipulated into:

$$d_t = \left\| \frac{\nabla \mathcal{L}(w_t)}{\|\nabla \mathcal{L}(w_t)\|} - \frac{\nabla \mathcal{L}(w_{t+1})}{\|\nabla \mathcal{L}(w_{t+1})\|} \right\| \leq 2. \quad (4)$$

Thus, the difference of updating $d_t$ under UGD for any $w$ at step $t \to t+1$ is uniformly bounded.

∎

### B. Proof of Theorem 2

**Theorem 2**. *For any distribution $\mathcal{D}$, with probability $1 - \delta$ over the choice of the training set $\mathcal{S} \sim \mathcal{D}$,*

$$\mathcal{L}_{\mathcal{D}} \leq \max_{\|\epsilon\|=1} \mathcal{L}_{\mathcal{S}}(w + \epsilon) \\ + \sqrt{\frac{k \log\left(1 + \|w\|_2^2 \left(1 + \sqrt{\frac{\log(n)}{k}}\right)^2\right) + 4 \log \frac{n}{\delta} + \tilde{\mathcal{O}}(1)}{n-1}}, \quad (5)$$

*where $n = |\mathcal{S}|$, $k$ is the number of parameters, and we assumed $\mathcal{L}_{\mathcal{D}}(w) \leq \mathbb{E}_{\epsilon_i \sim \mathcal{N}(0,\sigma^2)}[\mathcal{L}_{\mathcal{D}}(w + \epsilon)]$.*

Based on [37] (Theorem 3.2), the generalization of a perturbated module is bounded by

$$\mathbb{E}_{\epsilon_i \sim \mathcal{N}(0,\sigma^2)}[\mathcal{L}_{\mathcal{D}}(w + \epsilon)] \quad (6)$$

$$\leq \mathbb{E}_{\epsilon_i \sim \mathcal{N}(0,\sigma^2)}[\mathcal{L}_{\mathcal{S}}(w + \epsilon)] + \\ \sqrt{\frac{k \log\left(1 + \frac{\|w\|_2^2}{k\sigma^2}\left(1 + \sqrt{\frac{\log(n)}{k}}\right)^2\right) + 4 \log \frac{n}{\delta} + \tilde{\mathcal{O}}(1)}{n-1}}.$$

In the case of perturbation in UGD, the generalization bound follows $\|\epsilon\| = 1$ from the unit gradient, so there is a probability of $1 - 1/\sqrt{n}$ that the generalization can be bounded by

$$\mathcal{L}_{\mathcal{D}}(w) \leq \left(1 - \frac{1}{\sqrt{n}}\right) \max_{\|\epsilon\|=1} \mathcal{L}_{\mathcal{S}}(w + \epsilon) + \frac{1}{\sqrt{n}} \\ + \sqrt{\frac{k \log(1 + \|w\|_2^2\left(1 + \sqrt{\frac{\log(n)}{k}}\right)^2) + 4 \log \frac{n}{\delta} + \tilde{\mathcal{O}}(1)}{n-1}} \\ \leq \max_{\|\epsilon\|=1} \mathcal{L}_{\mathcal{S}}(w + \epsilon) \\ + \sqrt{\frac{k \log(1 + \|w\|_2^2\left(1 + \sqrt{\frac{\log(n)}{k}}\right)^2) + 4 \log \frac{n}{\delta} + \tilde{\mathcal{O}}(1)}{n-1}}. \quad (1)$$

∎

### C. Tables of Evaluation in end-to-end training

TABLE IV. CIFAR-10 END-TO-END TRAINING RESULTS (TOP-1%)

| Model / Optimizer | VGG-16 | ResNet-18 | DenseNet-121* | UPANet-16 | Overall Average |
|---|---|---|---|---|---|
| SGD | 93.99±0.19 | 95.60±0.08 | 94.81±0.14 | 95.03±0.14 | 94.86 |
| Adagrad | 91.11±0.02 | 90.49±0.15 | 89.03±0.02 | 91.57±0.03 | 90.55 |
| UGD (NGD-FM) | 94.55±0.14 | 95.66±0.14 | 94.84±0.13 | 94.87±0.06 | 94.98 |
| NGD-CW | 94.31±0.14 | 95.51±0.12 | 94.84±0.10 | 94.93±0.09 | 94.90 |
| AMP | 89.28±0.12 | 93.09±0.13 | 92.30±0.01 | 93.17±0.20 | 91.96 |
| SAM | 94.13±0.17 | 95.57±0.04 | 94.74±0.12 | 95.21±0.17 | 94.91 |
| ASAM | 94.05±0.00 | 95.16±0.16 | 94.19±0.16 | 94.67±0.04 | 94.52 |
| PUGD | **94.60±0.06** | **95.91±0.01** | **95.03±0.02** | **95.41±0.02** | **95.24** |

* DenseNet-121: growth rate in 16.

TABLE V. CIFAR-100 END-TO-END TRAINING RESULTS (TOP-1%)

| Model / Optimizer | VGG-16 | ResNet-18 | DenseNet-121* | UPANet-16 | Overall Average |
|---|---|---|---|---|---|
| SGD | 74.18±0.01 | 78.42±0.02 | 76.08±0.08 | 76.07±0.13 | 76.19 |
| Adagrad | 60.68±0.04 | 64.85±0.10 | 61.21±0.03 | 68.22±0.15 | 63.74 |
| UGD (NGD-FM) | 75.69±0.02 | 79.62±0.05 | 75.11±0.20 | 75.03±0.07 | 76.63 |
| NGD-CW | 76.04±0.08 | 79.79±0.06 | 75.53±0.07 | 75.36±0.12 | 76.68 |
| AMP | 52.66±0.14 | 71.62±0.12 | 66.81±0.02 | 71.39±0.12 | 76.70 |
| SAM | 74.12±0.16 | 77.73±0.19 | 76.90±0.18 | 76.70±0.20 | 76.36 |
| ASAM | 73.52±0.18 | 76.99±0.02 | 76.63±0.11 | 75.07±0.04 | 75.55 |
| PUGD | **76.77±0.09** | **80.46±0.07** | **77.77±0.04** | **78.19±0.06** | **78.30** |

* DenseNet-121: growth rate in 16.